\documentclass[a4paper,fleqn]{cas-dc}

\usepackage[authoryear]{natbib}

\usepackage{multirow}%
\usepackage{amsmath,amssymb,amsfonts}
\usepackage{mathrsfs}%
\usepackage{listings}%
\usepackage{graphicx}
\usepackage{adjustbox}
\usepackage{pifont}
\usepackage{booktabs}
\usepackage{algorithm,algorithmic}
\usepackage{pifont}
\usepackage{tabularx}
\usepackage{hyperref}
\usepackage{bbding}
\usepackage[normalem]{ulem}
\useunder{\uline}{\ul}{}
\usepackage{makecell}
\usepackage{booktabs}   

\definecolor{mygray}{RGB}{239, 239, 239}
\definecolor{darkgreen}{rgb}{0.0, 0.5, 0.0}
\definecolor{darkred}{rgb}{0.55, 0.0, 0.0}

\def\tsc#1{\csdef{#1}{\textsc{\lowercase{#1}}\xspace}}
\tsc{WGM}
\tsc{QE}

\begin{document}
\let\WriteBookmarks\relax
\def\floatpagepagefraction{1}
\def\textpagefraction{.001}

\shorttitle{}    

\shortauthors{}  

\title [mode = title]{VisionCAD: An Integration-Free Radiology Copilot Framework}



%

\author[1]{Jiaming Li}


\fnmark[1]




\affiliation[1]{organization={School of Biomedical Engineering \& State Key Laboratory of Advanced Medical Materials and Devices, ShanghaiTech University},
            city={Shanghai},
            postcode={201210}, 
            state={Shanghai},
            country={China}}
\affiliation[2]{organization={Shanghai United Imaging Intelligence Co. Ltd.},
            city={Shanghai},
            postcode={200230}, 
            state={Shanghai},
            country={China}}
\affiliation[3]{organization={Shanghai Clinical Research and Trial Center},
            city={Shanghai},
            postcode={201210}, 
            state={Shanghai},
            country={China}}

\author[1]{Junlei Wu}

\fnmark[1]

\author[1]{Sheng Wang}
\fnmark[1]

\author[1]{Honglin Xiong}

\author[1]{Jiangdong Cai}

\author[1]{Zihao Zhao}

\author[1]{Yitao Zhu}

\author[1]{Yuan Yin}

\author[1,2,3]{Dinggang Shen}[orcid=0000-0002-7934-5698]
\cormark[1]
\ead{dgshen@shanghaitech.edu.cn}

\author[1,3]{Qian Wang}[orcid=0000-0002-3490-3836]
\cormark[1]
\ead{qianwang@shanghaitech.edu.cn}




\cortext[1]{Corresponding author}

\fntext[1]{These authors contributed equally to this work.}


\begin{abstract}
    Widespread clinical deployment of computer-aided diagnosis (CAD) systems is hindered by the challenge of integrating with existing hospital IT infrastructure. Here, we introduce VisionCAD, a vision-based radiological assistance framework that circumvents this barrier by capturing medical images directly from displays using a camera system. The framework operates through an automated pipeline that detects, restores, and analyzes on-screen medical images, transforming camera-captured visual data into diagnostic-quality images suitable for automated analysis and report generation. We validated VisionCAD across diverse medical imaging datasets, demonstrating that our modular architecture can flexibly utilize state-of-the-art diagnostic models for specific tasks. The system achieves diagnostic performance comparable to conventional CAD systems operating on original digital images, with an F1-score degradation typically less than 2\% across classification tasks, while natural language generation metrics for automated reports remain within 1\% of those derived from original images. By requiring only a camera device and standard computing resources, VisionCAD offers an accessible approach for AI-assisted diagnosis, enabling the deployment of diagnostic capabilities in diverse clinical settings without modifications to existing infrastructure.
\end{abstract}

\begin{highlights}
    \item \textbf{Integration-Free AI Diagnostic Framework}: VisionCAD eliminates hospital IT integration requirements by capturing medical images directly from displays using camera-based technology, enabling rapid deployment across diverse clinical settings.
    
    \item \textbf{Modular Architecture with SOTA Model Integration}: The framework flexibly integrates modality-specific state-of-the-art models (e.g., Ark$^{+}$ for chest X-rays, fine-tuned ViT for specialized tasks), ensuring optimal diagnostic performance for each imaging modality.
    
    \item \textbf{Minimal Performance Degradation}: Comprehensive validation across four medical imaging datasets demonstrates diagnostic performance with F1-score degradation within 2\% compared to systems operating on original digital images, with automated report generation metrics remaining within 1\% of original-image baselines.
    \end{highlights}


\maketitle


\section{Introduction}
\label{sec1}

Computer-aided diagnosis (CAD) systems have demonstrated substantial capabilities in medical image analysis, achieving accuracy rates that often match or exceed human expert performance across various diagnostic tasks~\citep{zhao2024mining, wang2024interactive, wang2025learning}. However, despite these technological advances, widespread clinical adoption of CAD systems remains limited, with integration barriers representing one of the most significant obstacles to their deployment in real-world healthcare settings.

A primary challenge lies in the complex integration requirements with existing hospital information systems (HIS). Modern healthcare facilities operate sophisticated interconnected networks comprising specialized subsystems including radiology information systems (RIS), picture archiving and communication systems (PACS), electronic health records (EHR), and laboratory information systems (LIS). Incorporating CAD tools into this ecosystem demands extensive customization, rigorous compatibility testing, and substantial infrastructure modifications. This complexity is compounded when multiple CAD algorithms from different vendors must operate within the same clinical environment. Furthermore, the heterogeneous nature of hospital IT systems across institutions ensures that these integration challenges persist.

\begin{figure}
    \centering
    \includegraphics[width=1.0\linewidth]{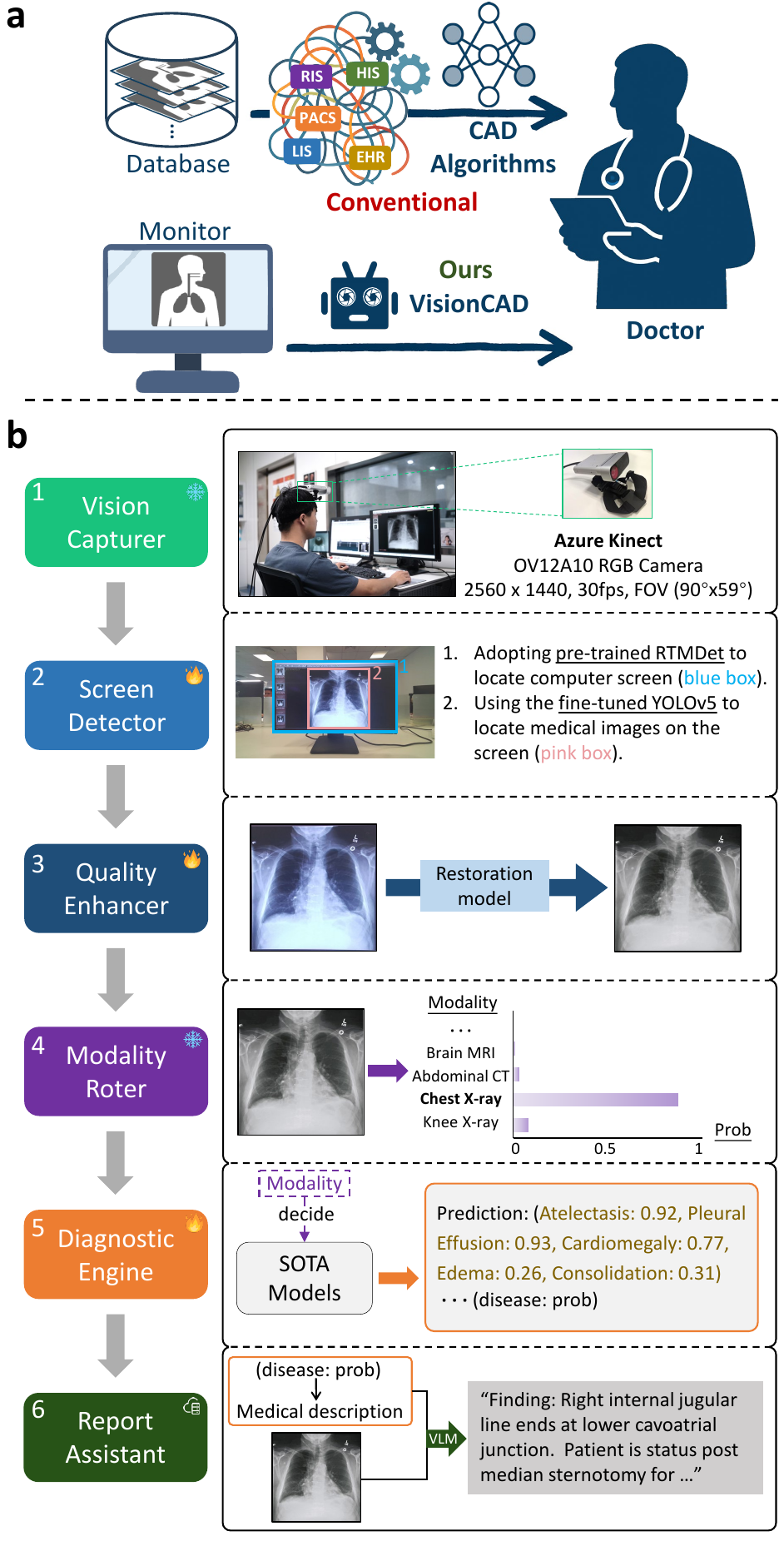}
    \caption{Integration paradigms for radiological diagnosis. (\textbf{a}) Conventional CAD systems require complex integration with hospital IT infrastructure, while VisionCAD bypasses these challenges through direct image capture.
    (\textbf{b}) The VisionCAD workflow proceeds through six components: (1) Vision Capturer captures images from displays; 
    (2) Screen Detector isolates medical image; 
    (3) Quality Enhancer restores image quality; 
    (4) Modality Router identifies image type; 
    (5) Diagnostic Engine performs analysis; 
    (6) Report Assistant generates clinical reports.
    }
    \label{fig:pipeline}
\end{figure}

Beyond technical barriers, clinical workflow integration presents additional challenges. Conventional CAD systems often require radiologists to switch between multiple interfaces, disrupting their examination flow and potentially introducing inefficiencies. The need for specialized workstations, dedicated displays, and custom software installations further complicates deployment, particularly in resource-constrained environments. These challenges have created a gap between the demonstrated potential of CAD technology and its practical implementation in clinical practice, highlighting the need for an alternative paradigm that circumvents these IT and workflow integration hurdles entirely.

Recent advances in camera technology~\citep{goesele2025imaging, bipat2019analyzing, engel2023project} offer new possibilities for clinical AI deployment. Driven by progress in computational photography, low-power processors, high-resolution sensors, and image processing algorithms, camera-based systems have evolved into platforms suitable for professional applications. Examples like Be-My-Eyes\footnote{\href{https://openai.com/index/be-my-eyes/}{https://openai.com/be-my-eyes/}} demonstrate how vision AI can assist with complex real-world visual tasks. In healthcare, vision-based technologies offer advantages that include real-time processing, workflow integration, and remote consultation support without requiring complex hardware modifications~\citep{marvasti2024unlocking, parkar2025next}. Camera-based platforms provide a practical approach toward scalable medical image analysis by aligning technical capabilities with the need for seamless CAD integration.

In this work, we introduce VisionCAD, a framework that leverages camera-based technology to provide AI diagnostic support. As illustrated in Figure~\ref{fig:pipeline}(a), traditional CAD systems require complex integration with multiple hospital information subsystems, creating significant deployment barriers. VisionCAD addresses this challenge by employing a camera that captures medical images directly from radiological displays during routine diagnostic workflows. This approach effectively bypasses traditional IT infrastructure requirements while preserving existing clinical workflows.

VisionCAD processes visual information through a six-component pipeline. The \textit{Vision Capturer} acquires images from displays. The \textit{Screen Detector} isolates medical content from surrounding interface elements. The \textit{Quality Enhancer} applies restoration algorithms to correct capture-related degradations. The \textit{Modality Router} identifies the image type to select the appropriate analysis pathway. The \textit{Diagnostic Engine} applies state-of-the-art models for specific clinical tasks. Finally, the \textit{Report Assistant} synthesizes findings into structured clinical reports. We validated our approach through comprehensive evaluations across multiple medical imaging datasets, demonstrating diagnostic performance comparable to conventional CAD systems while maintaining minimal deployment requirements.

Our primary contributions are threefold:

\begin{itemize}
    \item We propose VisionCAD, a camera-based diagnostic pipeline that enables AI-assisted radiology without hospital IT integration.
    \item We design a modular framework adaptable to diverse medical image diagnosis tasks using state-of-the-art models.
    \item We validate VisionCAD across multiple datasets, demonstrating near-parity with conventional CAD systems in diagnostic accuracy and report quality.
\end{itemize}


\section{Related Work}

VisionCAD integrates two key technological domains: image restoration techniques for processing degraded radiographs and pre-trained foundation models for efficient medical image analysis. While each domain has evolved independently, their integration enables robust diagnostic performance from camera-captured inputs, bridging the gap between real-world image acquisition and AI-assisted interpretation.
The following sections examine how each domain contributes to our framework within the broader landscape of AI-assisted medical diagnostics.

\subsection{Advances in Processing Degraded Images}

The deployment of radiographic AI beyond native PACS environments necessitates model robustness to images captured by commodity cameras. This capture process introduces a host of artifacts not present in original digital images, such as perspective distortions, variable illumination, glare, and moiré patterns, all of which can significantly impair diagnostic model performance.

Several key studies have quantified the impact of such degradations. The CheXphoto~\citep{phillips2020chexphoto} dataset, for instance, demonstrated that both authentic smartphone captures and synthetic photographic transformations of chest X-rays can degrade algorithmic performance, highlighting the need for robust models in real-world deployment scenarios. Similarly, 
CheXwhatsApp~\citep{antony2025chexwhatsapp} revealed that even subtle artifacts from lossy compression, common in mobile health applications, can substantially reduce the reliability and explainability of state-of-the-art models. Extending this line of inquiry to other modalities, Quan et al.~\citep{quan2022recovering} tackled the challenge of restoring camera-captured CT films, developing the CTFilm20K dataset and the FIReNet framework to correct for geometric and illumination distortions. These works collectively establish that although camera-based capture presents substantial challenges, specialized processing can effectively recover diagnostic fidelity of medical images..

These restoration techniques often benefit from broader advances in computer vision. Early deep learning architectures like UNet~\citep{ronneberger2015unet} established the encoder-decoder paradigm that remains influential. More recently, specialized models have emerged, including MSEC~\citep{afifi2020msec} for exposure correction and GRL~\citep{li2023grl} for hierarchical restoration. The advent of Transformer architectures has further advanced the field, with models like UFormer~\citep{wang2022uformer}, SwinIR~\citep{liang2021swinir}, UHDFormer~\citep{wang2024uhdformer}, and Restormer~\citep{zamir2022restormer} setting strong benchmarks on natural image restoration tasks. For a system like VisionCAD, leveraging these powerful restoration backbones is essential to precondition the captured images, ensuring they are of sufficient quality for reliable downstream diagnostic analysis.

\subsection{Foundational AI Technologies for VisionCAD}
VisionCAD's modular architecture is built upon a suite of pre-trained foundation models, each selected to perform a specialized function within the diagnostic pipeline. The emergence of these models has provided new capabilities for medical image analysis, offering adaptable representations for diverse clinical tasks.
In the screen detection stage, real-time object detection is critical. High-efficiency architectures like RTMDet~\citep{lyu2022rtmdet} and YOLOv5~\citep{jocher2020ultralytics} provide the rapid and accurate localization needed to isolate medical imagery from complex user interfaces.
For modality routing, the foundational CLIP~\citep{radford2021learning} model pioneered joint visual and textual embeddings, leading to specialized variants like BiomedCLIP~\citep{zhang2024biomedclip} that are fine-tuned on biomedical corpora for improved domain-specific performance.

For the core diagnostic capability, recent advances in modular AI architectures have enabled more precise diagnostic classification by integrating task-specific models rather than relying on general-purpose solutions. For chest radiography, Ark$^{+}$~\citep{ma2025fully} has emerged as a state-of-the-art model, offering comprehensive interpretation capabilities across a wide range of thoracic findings. In domains where no specialized foundation models are available (such as osteoarthritis classification from knee X-rays) Vision Transformer (ViT) architectures can be fine-tuned on curated datasets to achieve high task-specific performance.

Parallel progress in automated report generation has been driven by the development of Vision Language Models (VLMs) and Multimodal Large Language Models (MLLMs). Systems such as MedPaLM~\citep{tu2024towards}, GPT-4~\citep{hurst2024gpt}, and Gemini~\citep{comanici2025gemini} have demonstrated strong capabilities in interpreting medical images and producing structured, clinically coherent radiology reports that align with established diagnostic conventions~\citep{hartsock2024vision}. These models collectively represent a shift toward more adaptable and scalable solutions for AI-assisted medical interpretation.

\section{Method}
The VisionCAD system, shown in Figure~\ref{fig:pipeline}(b), consists of six  components that work together to analyze and interpret medical images captured directly from a screen.
\begin{enumerate}
    \item \textbf{Vision Capturer} captures medical images directly from diagnostic displays using a camera positioned to simulate a radiologist's natural viewing perspective.
    \item \textbf{Screen Detector} identifies and isolates medical image content from surrounding interface elements using real-time object detection algorithms.
    \item \textbf{Quality Enhancer} employs specialized image restoration techniques to correct capture-induced distortions and enhance diagnostic image quality.
    \item \textbf{Modality Router} automatically determines the imaging modality (e.g., chest X-ray, knee X-ray) using pre-trained VLMs for zero-shot classification.
    \item \textbf{Diagnostic Engine} applies modality-specific models to perform diagnostic analysis, employing foundation models like Ark$^{+}$ for established tasks or fine-tuned models for specialized applications.
    \item \textbf{Report Assistant} generates structured clinical reports by integrating diagnostic results with multimodal language models, producing findings and impressions in a standard radiological format.
\end{enumerate}
The following sections examine each of these components in detail, describing their implementation, technical specifications, and contributions to the overall VisionCAD framework.

\subsection{Vision Capturer}
\label{sec:vision_capturer}
The Vision Capturer serves as the input mechanism for the VisionCAD system, responsible for accurately capturing medical images from displays.
We selected the Microsoft Azure Kinect for our implementation based on its technical specifications that align with our research requirements. The device features a 2560×1440 resolution RGB camera with a wide field of view (H: 90°, V: 59°), providing the visual fidelity and coverage required for radiological applications. Kinect enables automated, repeatable, an scalable acquisition processes essential for systematic evaluation. The automated acquisition capability allows unattended batch capture for extended experiments, eliminating operator variability and ensuring consistent capture parameters. The repeatable acquisition ensures stable capture conditions, which is essential for fair comparison between original and processed images during evaluation. The scalable acquisition capability enables the system to handle varying experimental scales, from single-case studies to comprehensive dataset evaluations.

\begin{figure}[htbp]
    \centering
    \includegraphics[width=1\linewidth]{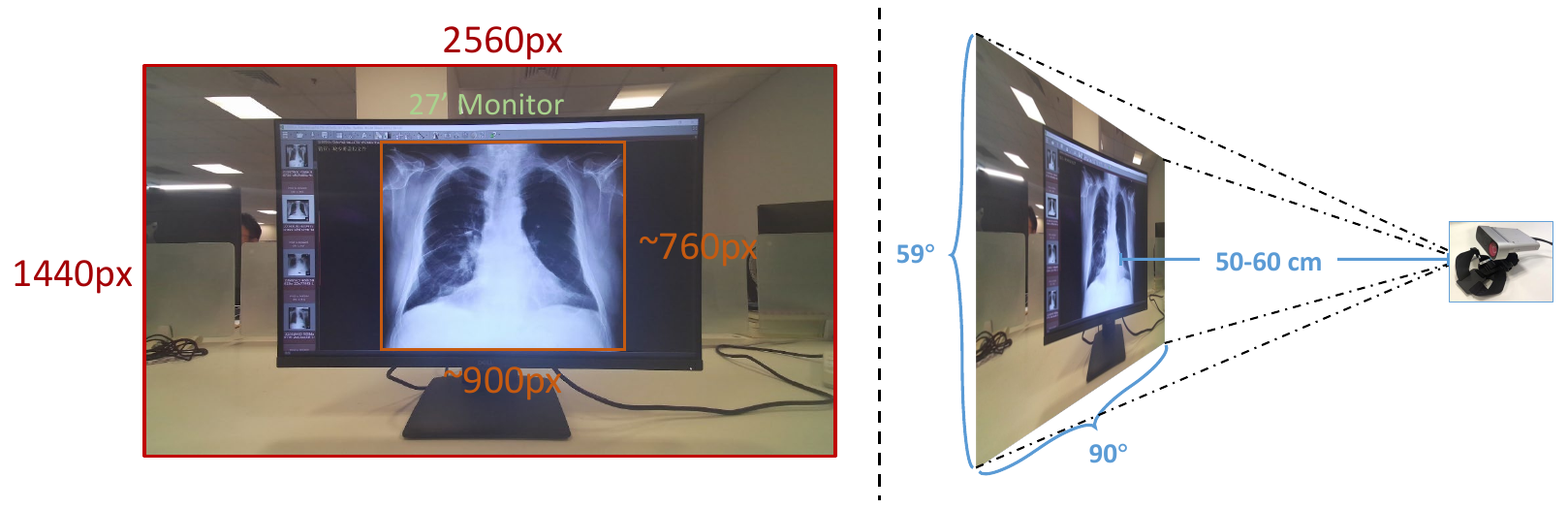}
    \caption{Setup of the Vision Capturer and its field of view characteristics. 
    The Kinect's wide-angle lens (Horizontal 90° Vertical 59°) effectively captures entire monitor displays from the standard radiologist viewing distance of 50–60 cm.}
    \label{fig:camera}
\end{figure}

Figure~\ref{fig:camera} illustrates the camera's field of view coverage and positioning relative to diagnostic displays. The camera captures entire monitor displays from the standard radiologist viewing distance of 50–60 cm, simulating the natural position and perspective of the radiologist.

For our prototype implementation, we employ a head-mounted setup that captures medical images directly from the radiologist's field of view. This configuration demonstrates the feasibility of camera-based diagnostic assistance in routine workflows. While effective for proof-of-concept validation, practical deployment envisions integration into lighter and more compact smart glasses that are equipped with built-in high quality cameras.

\subsection{Screen Detector}
\label{sec:screen_detector}
The Screen Detector locates medical images within the scene captured by the Vision Capturer. We propose a two-stage solution that first locates the monitor and then extracts the medical image from the system GUI. This approach consistently captures diagnostic screen content without being affected by viewing angles, lighting conditions, or display variations.

In the first stage, the Screen Detector identifies the monitor display within the captured field of view using a pre-trained Real-Time Multi-scale Detection (RTMDet) model~\citep{lyu2022rtmdet}. We employ the pre-trained model without additional fine-tuning, leveraging its feature extraction capabilities for rapid and accurate screen boundary detection. The model generates a segmentation mask that outlines the monitor display. From this irregular shape, we extract the closest quadrilateral approximation, which serves as the input for an affine transformation. This transformation normalizes the segmented screen region to a front-facing view, correcting for perspective distortion caused by the camera's angle and ensuring consistent performance across varying viewing positions.

\begin{figure}[htbp]
    \centering
    \includegraphics[width=\columnwidth]{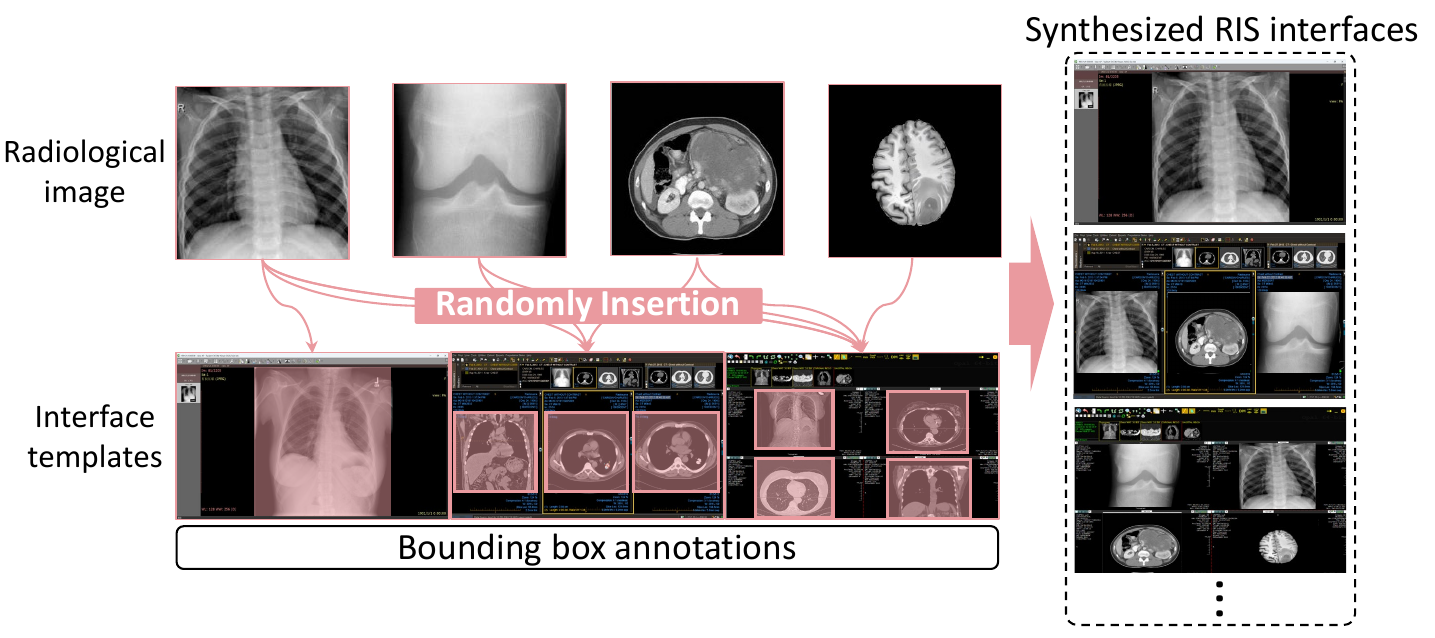}
    \caption{The dataset synthesis process for fine-tuning our medical image localization model. First, we collect 50 common radiology information system screenshots and manually annotate regions dedicated to displaying medical images. Next, we randomly insert diverse radiological images into these templates, synthesizing 10K training samples with bounding box annotations.}
    \label{fig:locator}
\end{figure}

The second stage focuses on locating the specific medical image content displayed on the screen interface. For this task, we implement a YOLOv5~\citep{jocher2020ultralytics} object detection architecture fine-tuned on our purpose-built dataset of radiological workstation interfaces. As illustrated in Figure~\ref{fig:locator}, we construct this specialized dataset by collecting screenshots of 50 diverse and representative RIS commonly used in clinical settings. We annotate regions dedicated to displaying medical images within these interfaces using bounding boxes. To generate sufficient training data, we randomly embed radiological images from various modalities into these annotated templates, synthesizing 10{,}000 training samples with ground-truth bounding box annotations. This synthetic dataset enables the fine-tuned YOLOv5 model to effectively differentiate medical images from surrounding interface elements such as toolbars, patient information panels, and measurement displays.

\subsection{Quality Enhancer}
The Quality Enhancer module addresses image degradation introduced during camera-based capture, which can compromise diagnostic accuracy. When a medical image displayed on a monitor is recorded by the Vision Capturer, the resulting image $I_{\text{captured}}$ differs from the original $I_{\text{original}}$ due to environmental factors including variable lighting, viewing angles, optical distortions, and display artifacts.
Our modular design supports seamless integration of state-of-the-art restoration architectures. We employ a unified restoration model trained to enhance image quality across diverse anatomical regions and imaging modalities. This shared model transforms degraded camera-captured inputs into high-fidelity outputs while preserving clinically relevant features, enabling consistent restoration performance without modality-specific retraining.

To train the restoration network, we construct a comprehensive paired dataset consisting of original medical images $I_{\text{original}}$ displayed on monitors and their corresponding captured versions $I_{\text{captured}}$ under varied conditions. The dataset includes chest X-rays, knee X-rays, brain MRI, abdominal MRI, and abdominal CT, totaling 22{,}975 2D image pairs. Following the procedure described in Section~\ref{sec:vision_capturer} and~\ref{sec:screen_detector}, we have systematically varied lighting (from bright overhead illumination to dim radiology reading room settings), viewing angles ($\pm$30° horizontal, $\pm$15° vertical), and camera parameters to simulate a wide range of real-world degradations.
Captured images are spatially aligned with their originals using the Scale-Invariant Feature Transform (SIFT) algorithm~\citep{sift2004lowe} to ensure pixel-level correspondence for ground-truth supervision.

After extensive benchmarking across multiple datasets and evaluation metrics (detailed in Section~\ref{sec:exp_enhancer}), we select Restormer~\citep{zamir2022restormer} as the backbone for our Quality Enhancer. It offers a favorable balance between restoration quality and computational efficiency, making it well-suited for clinical deployment.

\subsection{Modality Router}
To ensure optimal diagnostic performance across diverse medical imaging modalities, we use a modality recognition system that intelligently routes VisionCAD images to specialized analytical models.
Our approach leverages BiomedCLIP~\citep{zhang2024biomedclip}, a pre-trained VLM specifically designed for biomedical applications, which encodes both image and text into a unified embedding space. The classification process operates through a zero-shot mechanism: the system encodes the VisionCAD medical image and computes cosine similarities between this embedding and the embeddings of text descriptions representing various medical modalities (e.g., "chest X-ray", "knee X-ray", "brain MRI").

The zero-shot classification strategy achieves high accuracy across diverse imaging types while maintaining computational efficiency. Once the specific modality is determined, VisionCAD activates the corresponding specialized Diagnostic Engine, ensuring each medical image receives analysis from the most appropriate model available.

\subsection{Diagnostic Engine}
The Diagnostic Engine employs a task-specific model selection strategy that prioritizes the use of state-of-the-art (SOTA) models designed for particular imaging modalities and diagnostic tasks. This approach ensures optimal diagnostic performance by leveraging specialized models that have been specifically developed and validated for each clinical scenario.

For chest X-ray analysis, we integrate Ark$^{+}$~\citep{ma2025fully}, a foundation model that represents the current SOTA for comprehensive chest radiography interpretation. This model has been specifically trained on large-scale chest X-ray datasets and demonstrates superior performance in detecting multiple thoracic pathologies simultaneously. Its understanding of chest anatomy and pathology makes it suitable for complex multi-label chest imaging scenarios, such as those encountered in the MIMIC-CXR dataset.

For knee X-ray analysis in osteoarthritis classification, no specialized SOTA foundation model currently exists that matches the performance standards achieved in chest radiography. In such cases, we adopt a systematic approach: we fine-tune a Vision Transformer (ViT)~\citep{dosovitskiy2020image} architecture on the specific dataset to create a task-optimized model. The ViT backbone provides strong representational capabilities that can be effectively adapted to knee pathology assessment through targeted fine-tuning on the OAI dataset. This approach ensures consistent high performance even in domains where specialized foundation models are not yet available.

The selection process operates through the Modality Router, which identifies the imaging modality and automatically activates the corresponding diagnostic model. 
This strategy represents a pragmatic approach to clinical AI deployment, combining the advantages of foundation models where available with reliable custom-trained solutions where needed. As new SOTA models emerge for different imaging modalities, they can be readily integrated into the framework, ensuring that VisionCAD continuously benefits from advances in medical AI while maintaining consistent diagnostic quality across all supported imaging types.

\subsection{Report Assistant}
The Report Assistant module completes VisionCAD's diagnostic pipeline by transforming analytical outputs into comprehensive clinical reports. Our implementation employs a multimodal VLM to convert diagnostic findings into clinically relevant reports that follow established radiological conventions. The report generation process operates through two complementary mechanisms that prioritize both diagnostic accuracy and clinical utility.

First, we implement a structured post-processing approach that transforms numerical outputs from the Diagnostic Engine into standardized clinical terminology~\citep{wang2024interactive,zhao2024chatcad+}. As shown in Table~\ref{tab:prob2text}, prediction probabilities are systematically mapped to standardized clinical statements, enhancing interpretability for both VLMs and medical professionals. This probability-to-text conversion establishes a consistent semantic framework for expressing diagnostic confidence levels across different conditions and imaging modalities, ensuring that quantitative predictions maintain their clinical significance throughout the reporting pipeline.

Second, we feed the VisionCAD-processed image and standardized findings into a multimodal VLM using a structured prompt. This prompt concatenates (a) the probability-to-text list of disease findings produced by our classifier with (b) a fixed instruction asking the model to return a radiology report with ``Findings'' and ``Impression'' sections in the format ``Findings: \{\} Impression: \{\}''. The VLM then generates the final report for the target image following this structure. For all experiments, we set the decoding temperature to 0.2 to ensure deterministic and consistent outputs, with a maximum token limit of 1024 to accommodate detailed reports.

For chest X-ray analysis, we demonstrate this template on MIMIC-CXR test cases; the resulting outputs exhibit standard clinical phrasing in the ``Findings'' section and concise bullet-style items in ``Impression''. The generated reports adhere to established clinical documentation practices, providing findings, impressions, and recommendations in a format familiar to medical professionals.

\begin{table}[htbp]
    \centering
        \caption{Probability-to-text conversion mapping used in the Report Assistant module. 
        Each predicted probability range corresponds to a specific descriptive statement template, where the term \{\textbf{disease}\} is replaced with the relevant medical condition detected by the Diagnostic Engine.
        }
        \label{tab:prob2text}
    \begin{tabular}{lc}
    \toprule
    Predicted Probability      & Textual Description\\
    \hline
    [0, 0.2)   & No sign of \{\textbf{disease}\} \\
    {[}0.2, 0.5) & Small possibilty of \{\textbf{disease}\}                             \\
    {[}0.5, 0.9) & Likely to have \{\textbf{disease}\}                            \\
    {[}0.9, 1{]} & Definitely have \{\textbf{disease}\}             \\
    \bottomrule
    \end{tabular}
\end{table}

\section{Experimental Results}
\label{sec:exp}

\subsection{Datasets and Implementations}
\label{sec:data}

To comprehensively evaluate whether integration-free VisionCAD delivers diagnostic performance comparable to conventional CAD systems that operate on original digital images, we employ diverse medical imaging datasets with varying complexity levels and clinical characteristics.

\begin{itemize}
    \item \textbf{PneumoniaMNIST}~\citep{yang2023medmnist}: A pediatric chest X-ray dataset for binary pneumonia classification containing 5,856 images (224×224 pixels) from the MedMNIST collection. Following the official split, we use 4,708 images for training (for the Quality Enhancer and Diagnostic Engine), 524 images for validation, and 624 images for testing.

    \item \textbf{OAI}~\citep{chen2019fully}: A knee X-ray dataset from the Osteoarthritis Initiative comprising 5,778 images from 2,276 participants. It features 5 distinct stages of osteoarthritis severity based on the Kellgren and Lawrence (KL) grading system. We adopt a custom 8:1:1 split, yielding 5,200 training images and 578 test images; the training set is used for the Quality Enhancer and Diagnostic Engine, the validation set for model selection, and the test set for evaluation.

    \item \textbf{Nodule}~\citep{shen2023image}: A frontal-view chest X-ray dataset comprising 6,048 clinically acquired high-resolution images collected from our collaborative hospitals, with 3,024 images diagnosed as containing pulmonary nodules and 3,024 normal cases. We randomly select 605 images from each category for testing (1,210 total), with the remaining 4,838 images used for training.

    \item \textbf{MIMIC-CXR}~\citep{johnson2019mimic}: A large-scale dataset of 377,110 chest X-ray images with associated free-text reports. For our report generation evaluation, we selected 500 cases (811 images) from the official hold-out test set, covering all 14 standard CheXpert pathology labels. For training the Quality Enhancer, we sample 5,000 images from the official training split, consisting of 4,500 for training and 500 for validation. This dataset serves as a standard benchmark for validating VisionCAD's performance in complex, real-world clinical scenarios.
\end{itemize}

To replicate realistic radiological diagnosis environments, we configure our experimental system with the following components:
\begin{itemize}
    \item \textbf{Display Monitor}: A consumer-grade Dell monitor (D2720DS) with default settings, providing 2560×1440 resolution to represent typical medical displays.

    \item \textbf{Vision Device}: The Microsoft Azure Kinect, featuring a RGB camera (2560×1440 at 30fps) and a wide field-of-view (H: 90°, V: 59°), chosen for its high-fidelity data acquisition and robust performance in varied lighting conditions.

    \item \textbf{Vision Language Model}: SOTA models including GPT-4.1 and Gemini-2.5-pro, selected to evaluate the framework's compatibility across different VLM paradigms. For all report generation experiments, we set the temperature to 0.2 for deterministic outputs and a maximum token limit of 1024.
\end{itemize}

\subsection{Visual Quality of VisionCAD Processing}
\label{sec:exp_enhancer}

\begin{figure*}[!h]
    \centering
    \includegraphics[width=\linewidth]{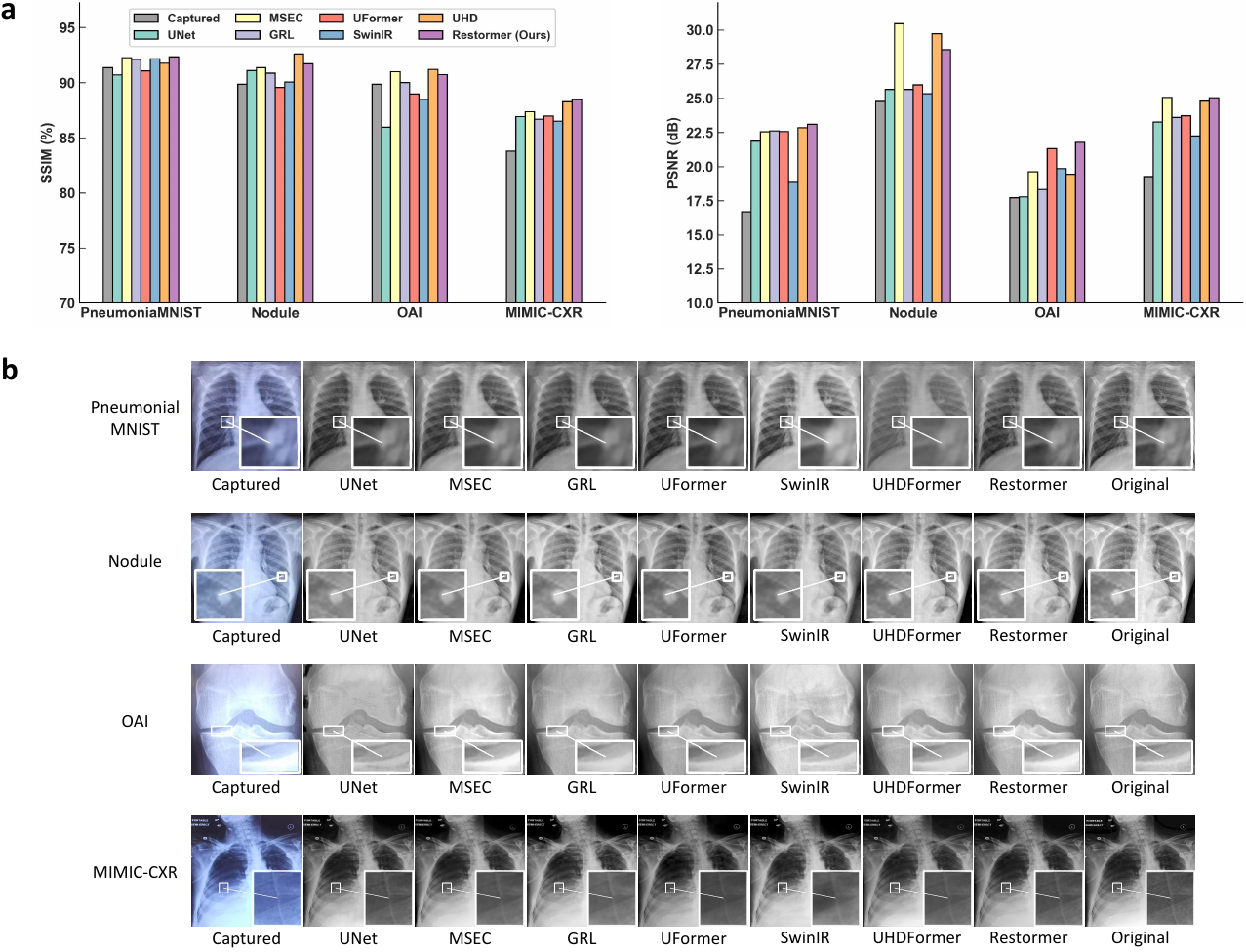}
    \caption{Quality Enhancer performance evaluation across different restoration methods. (\textbf{a}) We evaluated seven restoration methods (UNet, MSEC, GRL, UFormer, SwinIR, UHDFormer, Restormer) on four medical imaging datasets (PneumoniaMNIST, Nodule, OAI, MIMIC-CXR). Image quality is assessed using PSNR and SSIM metrics (SSIM values multiplied by 100 for display consistency). (\textbf{b}) Visual comparisons illustrate how different methods correct capture-related degradations. Within each row, images progress from Captured (raw detector output) through the seven restoration methods to Original (ground truth). Zoomed-in regions support detailed inspection of restoration effectiveness.}
    \label{fig:visualization}
\end{figure*}

To assess image quality within the VisionCAD pipeline, we perform a dedicated evaluation at the output of the Quality Enhancer stage. This mid-pipeline checkpoint allows us to measure restoration performance before diagnostic inference, ensuring that downstream modules receive visually reliable inputs.

We compare seven representative restoration methods spanning diverse architectural paradigms. These include classical encoder-decoder models such as UNet~\citep{ronneberger2015unet}, exposure correc  tion techniques like MSEC~\citep{afifi2020msec}, hierarchical designs such as GRL~\citep{li2023grl}, and Transformer-based approaches including UFormer, SwinIR, UHDFormer, and Restormer~\citep{wang2022uformer, liang2021swinir, wang2024uhdformer, zamir2022restormer}, which leverage attention mechanisms for advanced feature reconstruction.
All methods are evaluated on the four datasets introduced in Section~\ref{sec:data}. 
Test sets are strictly excluded from restoration training to ensure fair comparison.

Quantitative results (Figure~\ref{fig:visualization}(a) and Table~\ref{tab:devices_restore}) show consistent gains over raw captured images. Performance varies notably across datasets, reflecting differences in initial capture quality and image characteristics. Captured image PSNR ranges from 16.68 (PneumoniaMNIST) to 24.76 (Nodule), indicating variable degradation levels depending on factors such as display brightness, room lighting, camera positioning, and the inherent contrast properties of each modality. 

Restormer demonstrates the most balanced performance: it achieves top PSNR and SSIM on PneumoniaMNIST, best SSIM on MIMIC-CXR, and ranks competitively across all metrics and datasets. Figure~\ref{fig:visualization}(b) provides qualitative examples showing correction of lighting, perspective, and optical artifacts.
Based on its consistent top-tier performance and stability across diverse imaging conditions, we adopt Restormer as the default restoration backbone for subsequent diagnostic experiments.

\subsection{Diagnostic Engine Performance Evaluation}

\begin{figure}[htbp]
    \centering
    \includegraphics[width=1.0\linewidth]{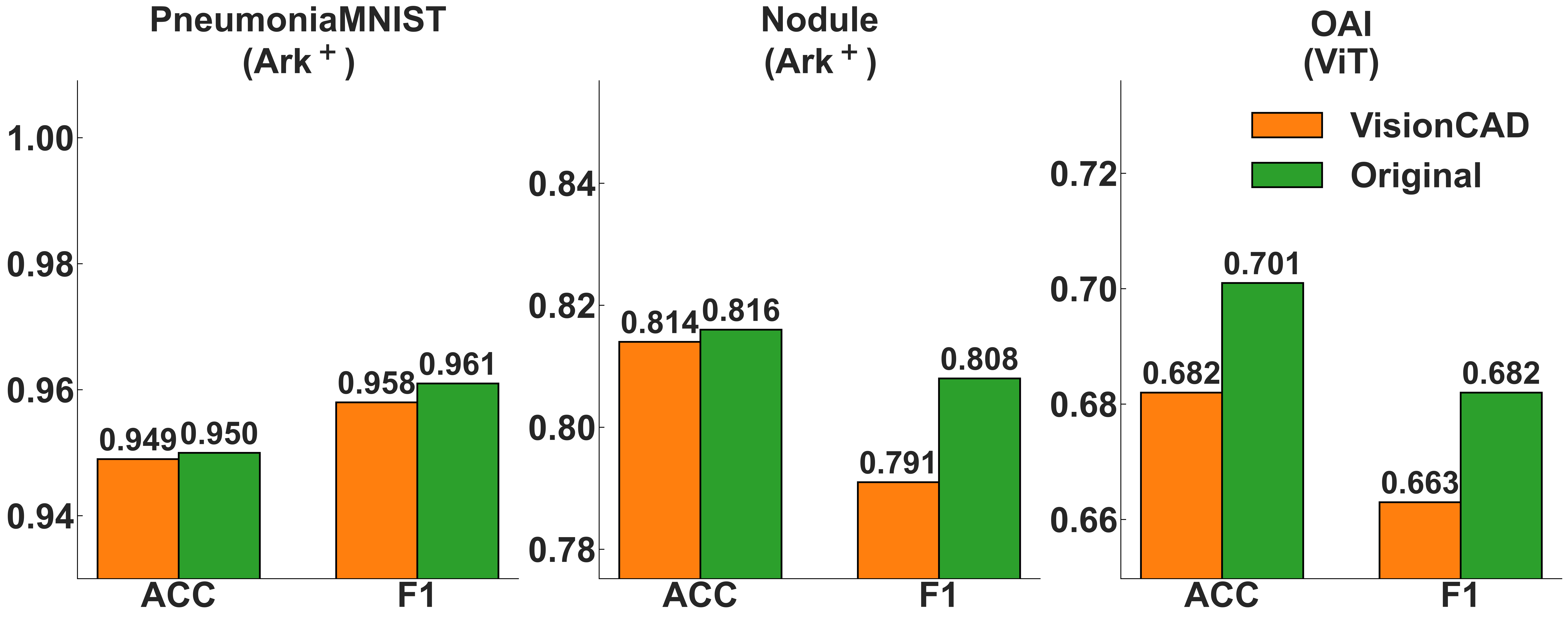}
    \caption{Performance evaluation using modality-specific SOTA models. Ark$^{+}$ represents the current SOTA for chest X-ray diagnosis, while ViT fine-tuning is employed for tasks without established foundation models. All models were trained on Original images and evaluated on both Original and VisionCAD test sets.}
    \label{fig:table2}
\end{figure}

\begin{figure*}[htbp]
    \centering
    \includegraphics[width=1.0\linewidth]{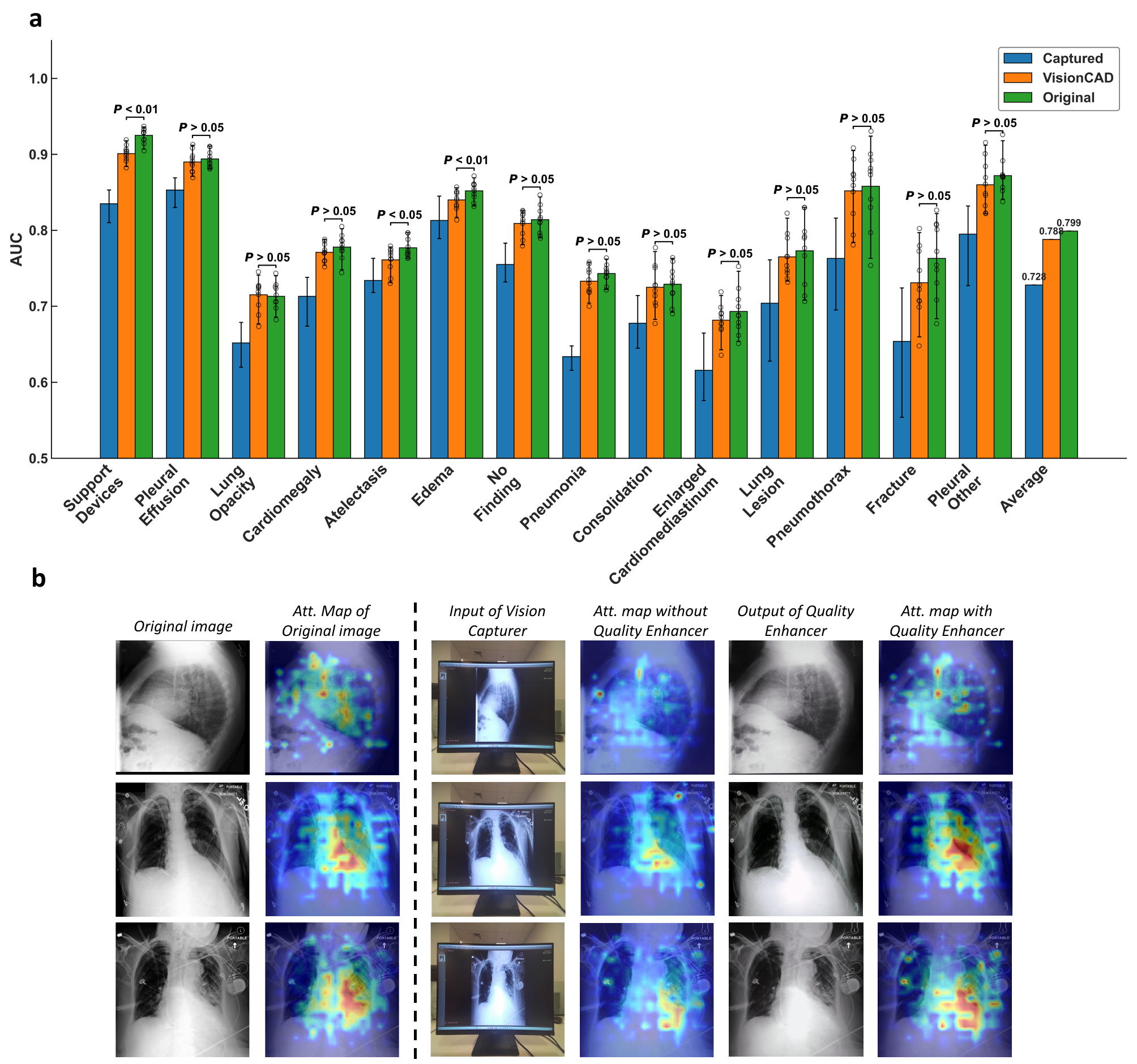}
    \caption{Performance evaluation of the Ark$^{+}$ diagnostic engine on the MIMIC-CXR dataset. (\textbf{a}) Comparison of Area Under the Curve (AUC) scores across 14 CheXpert conditions for three image types: Captured (raw camera images), VisionCAD (restored images), and Original. Each bar represents the mean AUC with 95\% confidence intervals derived from 1000 bootstrap samples. Statistical comparisons between VisionCAD and Original were performed using the Wilcoxon signed-rank test under the null hypothesis that there is no significant difference in diagnostic performance between the two image types. (\textbf{b}) Grad-CAM visualizations for a representative case, illustrating the model's attention on the ground-truth positive conditions. The attention maps correspond to three image stages: Original, VisionCAD-Detected, VisionCAD-Restored.}
    \label{fig:exp2}
\end{figure*}

A critical aspect of our evaluation is to validate whether VisionCAD, despite its non-disruptive capture method, can achieve diagnostic accuracy comparable to conventional CAD systems that have direct access to original medical images. Our evaluation strategy employs modality-specific SOTA models to ensure the most rigorous assessment possible for each imaging task.

\subsubsection{Single-Label Evaluation}

For chest X-ray analysis tasks (PneumoniaMNIST, Nodule detection), we employ Ark$^{+}$, which represents the current SOTA for comprehensive chest radiography interpretation. This foundation model has been specifically trained on large-scale chest X-ray datasets and demonstrates superior performance in detecting multiple thoracic pathologies. To adapt Ark$^{+}$ to these specific tasks, we have employed linear probing, where the pre-trained feature representations are frozen and only a linear classification head is trained on the target datasets. 
Since Ark$^{+}$ accepts 768×768 pixel input images, we resize both PneumoniaMNIST and Nodule dataset images to this resolution to ensure compatibility with the pre-trained model architecture. 
For knee X-ray analysis (OAI dataset), where no specialized SOTA foundation model exists, we fine-tuned a Vision Transformer (ViT) architecture on the specific dataset to ensure optimal performance for osteoarthritis classification. The ViT model operates at 224×224 pixel resolution, following standard practice for this architecture.

As shown in Figure~\ref{fig:table2}, all tests achieve strong baseline performance on original images. When evaluated on VisionCAD-processed images, they maintain high diagnostic accuracy with minimal performance degradation. For PneumoniaMNIST, Ark$^{+}$ with linear probing achieves an F1-score of 96.1\% on original images, with only a minor drop to 95.8\% on VisionCAD-processed images, demonstrating the effectiveness of our image restoration pipeline. For nodule detection, the F1-score shows a slight decrease from 80.8\% on original images to 79.1\% on VisionCAD-processed images, representing a degradation of 1.7 percentage points. The ViT model on the OAI dataset achieves an F1-score of 68.2\% on original images and 66.3\% on VisionCAD-processed images, representing a modest degradation of 1.9 percentage points.

These results demonstrate that our pipeline effectively preserves the diagnostic features necessary for high-performance analysis across different imaging modalities, whether using established foundation models with linear probing or custom-trained architectures. The consistent performance across all three datasets, with F1-score degradation ranging from 0.3\% to 1.9\%, validates the robustness of our approach.

\subsubsection{Multi-Label Evaluation}

To assess performance in complex clinical scenarios, we evaluated the MIMIC-CXR dataset for multi-label chest pathology detection. Figure~\ref{fig:exp2}(a) presents the detailed AUC scores across 14 conditions for three image types, with statistical confidence intervals derived from 1000 bootstrap samples. The results show that our Quality Enhancer module substantially improves diagnostic utility: the average AUC score improves from 0.728 for raw captured images to 0.788 for VisionCAD-processed images, nearly matching the 0.799 AUC achieved on original images.
The average performance difference between VisionCAD-processed and original images is only 0.011 AUC points, confirming the viability of our approach for real-world clinical deployment while leveraging SOTA diagnostic capabilities.

Figure~\ref{fig:exp2}(b) provides Grad-CAM visualizations demonstrating that the model's attention patterns on VisionCAD-processed images closely resemble those on original images. This indicates that our pipeline effectively preserves key pathological features, ensuring the diagnostic model focuses on clinically relevant regions even after image capture and restoration.

\subsection{Performance after Report Assistant}
\begin{figure*}[htbp]
    \centering
    \includegraphics[width=0.9\linewidth]{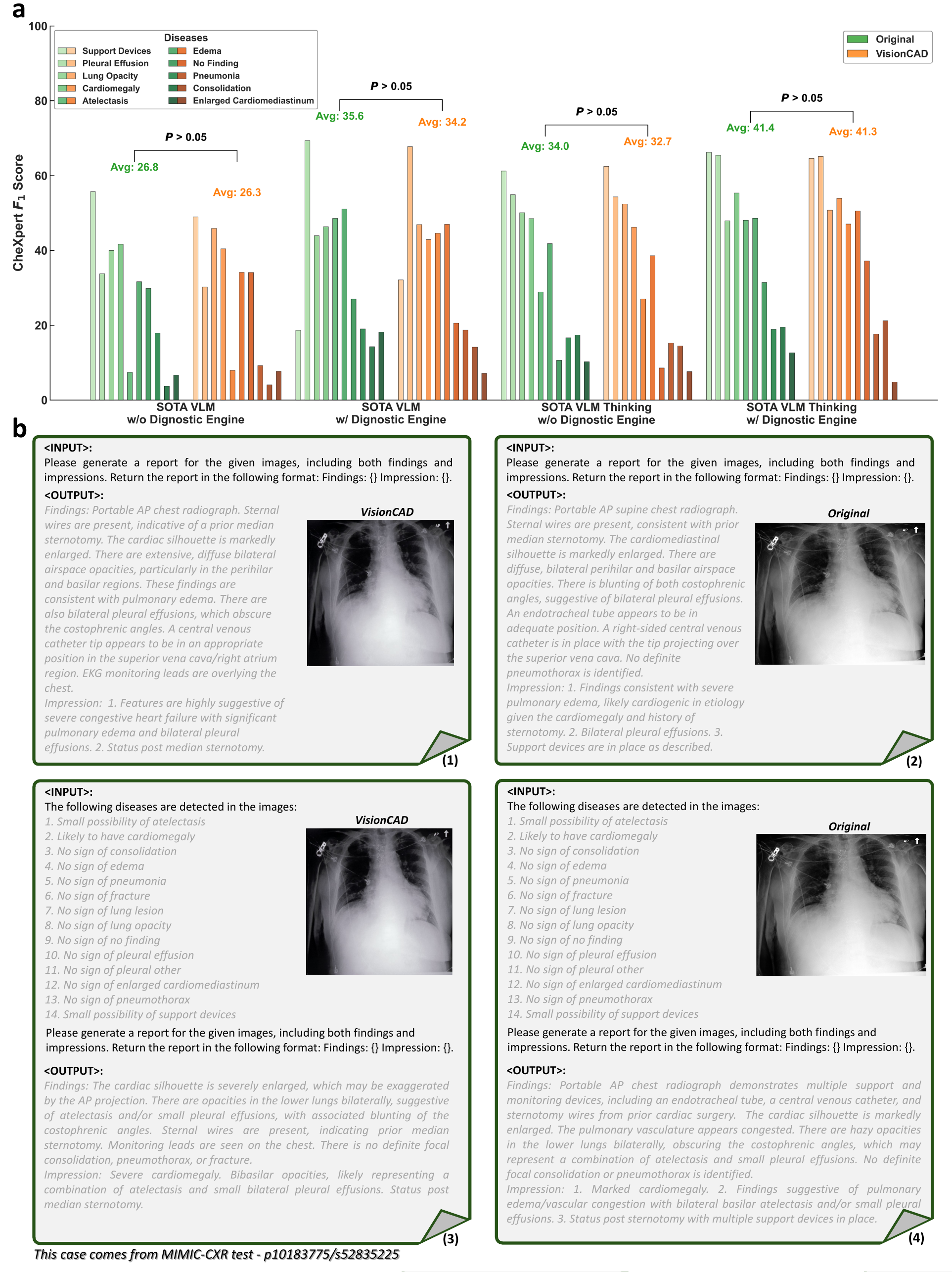}
    \caption{Evaluation of report generation performance on the MIMIC-CXR dataset. (\textbf{a}) Comparison of CheXpert $F_{1}$ scores across 10 common conditions for two model types: a direct-response VLM and a reasoning-based VLM. Performance is evaluated on both Original and VisionCAD images, with and without diagnostic priors from Ark$^{+}$. Each bar represents the CheXpert F$_{1}$ score for a condition, with overall average scores annotated above the respective groups. Wilcoxon signed-rank tests indicated no statistically significant difference between corresponding Original and VisionCAD pairs ($p$ > 0.05). (\textbf{b}) A representative case illustrating the report outputs generated by the reasoning-based VLM under four distinct settings: (1) VisionCAD w/o Ark$^{+}$, (2) Original w/o Ark$^{+}$, (3) VisionCAD w/ Ark$^{+}$, and (4) Original w/ Ark$^{+}$.}
    \label{fig:exp3}
\end{figure*}

\begin{table}[htbp]
    \renewcommand{\arraystretch}{1.1}
    \centering
    \caption{Evaluation of report generation quality across different image types and models. The table compares standard natural language generation metrics for reports generated from Original and VisionCAD-processed images, demonstrating the high fidelity of our restoration pipeline. SOTA VLM Model refers to GPT-4.1, and SOTA VLM thinking Model refers to Gemini-2.5-pro. All metric scores have been multiplied by 100 for clarity.}
    \label{tab:report_quality}
    \resizebox{\columnwidth}{!}{
    \begin{tabular}{l|cc|cc}
        \toprule
        \multirow{2}{*}{Metric} & \multicolumn{2}{c|}{SOTA VLM Model} & \multicolumn{2}{c}{SOTA VLM thinking Model} \\
        \cline{2-5}
        & Original & VisionCAD & Original & VisionCAD \\
        \hline
        BLEU-1   & 25.88 & 25.17 & 27.68 & 27.84 \\
        BLEU-2   & 13.56 & 12.84 & 15.64 & 15.64 \\
        BLEU-3   & 7.23  & 6.75  & 8.97  & 8.96  \\
        BLEU-4   & 3.84  & 3.52  & 5.22  & 5.31  \\
        ROUGE-L  & 17.07 & 16.55 & 18.82 & 18.98 \\
        METEOR   & 27.98 & 27.30 & 30.49 & 30.08 \\
        \bottomrule
    \end{tabular}
    }
\end{table}

To comprehensively evaluate VisionCAD's clinical report generation capabilities, we conduct experiments using the MIMIC-CXR dataset, which contains 377,110 chest X-ray images paired with professionally written radiological reports. We select a representative subset of 500 cases covering all CheXpert pathology labels to assess both diagnostic classification performance and natural language generation quality.

Our evaluation addresses two critical questions: (1) How does the inclusion of diagnostic classification results improve report generation accuracy? and (2) How effectively does our image restoration pipeline preserve diagnostic information for report generation? To ensure a robust and statistically significant evaluation, we focus our analysis on the 10 most prevalent pathological conditions in the MIMIC-CXR test set. These conditions collectively account for over 95\% of the positive findings in the hold-out test set, ensuring representative coverage while maintaining experimental feasibility.

For model comparison, we select GPT-4.1 as SOTA VLM model, representing the current state-of-the-art in direct response generation, and Gemini-2.5-pro as SOTA VLM thinking model, which incorporates advanced chain-of-thought capabilities for reasoning tasks.

\textbf{Diagnostic Accuracy Assessment.} We evaluat report generation performance using CheXpert $F_1$ scores across the 10 selected pathological conditions. Figure~\ref{fig:exp3}(a) presents the comparative results across both models. The evaluation considers four experimental conditions: original images with and without Ark$^{+}$ diagnostic input, and VisionCAD images with and without Ark$^{+}$ diagnostic input.

The results demonstrate the significant impact of incorporating diagnostic classification results into the report generation process. For the SOTA VLM model, the inclusion of Ark$^{+}$ diagnostic information consistently improves performance, with the average F1 score on original images increasing from 26.82 to 35.64. Similarly, the SOTA VLM thinking model achieves an even higher overall performance of 41.41 with diagnostic input on original images, representing a 21.6\% improvement over the baseline without diagnostic guidance (34.04).

When comparing original images to those processed by VisionCAD, the performance of both models remains remarkably close, especially when diagnostic priors are provided. For the SOTA VLM model with Ark$^{+}$ input, the average F1 score difference between original (35.64) and VisionCAD images (34.19) is only 1.45 points. The SOTA VLM thinking model shows even better resilience, with a negligible average decrease of just 0.12 points (41.41 vs 41.29), demonstrating robust performance across different image types. Statistical analysis using the Wilcoxon signed-rank test reveals no significant differences between original and VisionCAD-processed images within each diagnostic configuration ($p > 0.05$ for all paired comparisons), failing to reject the null hypothesis and supporting the conclusion that VisionCAD effectively preserves diagnostic information quality.

Analysis of specific pathological conditions reveals interesting patterns. Certain conditions show improved performance with VisionCAD processing, such as Lung Opacity, where both models consistently achieve higher F1 scores on VisionCAD images compared to original images. For instance, the SOTA VLM thinking model achieves 52.41 F1 score without Ark$^{+}$ on VisionCAD images versus 50.10 on original images, indicating that our restoration pipeline may enhance certain features beneficial for this condition. Conversely, conditions requiring detection of subtle features, such as Pneumonia and Enlarged Cardiomediastinum, show more variable performance, suggesting areas for future pipeline refinement.

\textbf{Linguistic Quality of Generated Reports.} To assess whether the image restoration process impacts the linguistic quality of the final output, we evaluat the generated reports using standard natural language generation (NLG) metrics, including BLEU~\citep{papineni2002bleu}, ROUGE-L~\citep{lin2004rouge}, METEOR~\citep{banerjee2005meteor}. Table~\ref{tab:report_quality} compares reports generated from VisionCAD-processed images against those from original images using both VLM types. Our analysis revealed that the Quality Enhancer pipeline effectively preserves the semantic content required for high-quality clinical reporting. For both VLM types, the NLG metrics for reports from VisionCAD-processed images are nearly identical to those from original images, with most scores differing by less than one percentage point.

This consistency across multiple metrics confirms that the visual information crucial for generating coherent and accurate clinical narratives is not compromised by our capture-and-restoration process. The findings suggest that VisionCAD can serve as a reliable source for automated report generation, producing outputs that are linguistically on par with those derived from original images.

\section{Conclusion and Discussion}

In this work, we present VisionCAD, a vision-based radiology assistance framework that addresses the challenge of integrating AI diagnostic models into clinical practice without requiring direct IT system integration. Through our modular approach of employing modality-specific SOTA models and systematic evaluation across multiple datasets, we have demonstrated that the system can deliver diagnostic support comparable to conventional CAD systems while maintaining workflow compatibility. Our experimental validation across foundational classification tasks and complex clinical scenarios confirms the viability of this approach for real-world deployment. 

VisionCAD presents a practical approach to integrating AI diagnostic tools into clinical practice through a non-disruptive, vision-based framework. By circumventing the persistent challenges of hospital IT integration, our approach offers an alternative pathway for bringing AI assistance to radiological workflows. Our evaluations demonstrate that VisionCAD achieves diagnostic performance comparable to models operating on original images while maintaining high fidelity in automated report generation.

A key strength of VisionCAD lies in its modular architecture that supports the integration of modality-specific SOTA models. Our strategy of employing Ark$^{+}$ for chest X-ray analysis and fine-tuned ViT models for tasks without established foundation models demonstrates the framework's adaptability. The minimal performance degradation observed between original and VisionCAD-processed images validates the effectiveness of our Quality Enhancer module in preserving clinically critical information. Our analysis of report generation reveals that incorporating structured diagnostic results from specialized models enhances the accuracy of reports generated by multimodal language models. The performance remains close when comparing original images to those processed by VisionCAD, indicating that our pipeline effectively preserves the semantic content required for high-quality clinical reporting. Beyond technical performance, VisionCAD offers practical advantages for broader deployment. Its independence from complex IT infrastructure and low hardware requirements make it accessible for diverse clinical settings.

VisionCAD has several limitations. First, system performance depends on vision device capabilities including battery life, processing power, and imaging quality. Second, certain radiological findings with subtle features exhibit more performance degradation, indicating a need for higher quality (less information loss) in image capture. Third, reliance on cloud-based models raises concerns about data privacy, security, and connectivity requirements. Finally, future real-world deployment will require more validation to address variable lighting conditions, diverse display types, and clinical workflow integration.

\printcredits

\bibliographystyle{cas-model2-names}

\bibliography{cas-refs}



\appendix
\section*{Appendix}
\label{sec:appendix}

This appendix provides detailed quantitative results supporting the main findings presented in the manuscript. Table~\ref{tab:devices_restore} presents comprehensive image quality metrics across seven restoration methods evaluated on four medical imaging datasets. Table~\ref{tab:Ark+_comparison} details the diagnostic classification performance using the Ark$^{+}$ engine on the MIMIC-CXR dataset with statistical significance testing. Table~\ref{tab:comparison_of_report_accurancy} reports automated report generation performance across different model configurations with and without diagnostic priors.

\begin{table*}[htbp]\footnotesize
    \centering
    \caption{Comparison of image quality metrics across different restoration methods using PneumoniaMNIST, Nodule, OAI, and MIMIC-CXR datasets. Results show PSNR and SSIM metrics for each restoration method. SSIM values are multiplied by 100 for display consistency with PSNR. Values are presented as mean±std from 5 independent runs. \textbf{Bold} indicates the best image quality result within each dataset. \underline{underline} indicates the second-best.}
    \label{tab:devices_restore}
        \begin{tabular}{c|cc|cc|cc|cc}
        \toprule
        \multirow{2}{*}{Model} & \multicolumn{2}{c|}{PneumoniaMNIST} & \multicolumn{2}{c|}{Nodule} & \multicolumn{2}{c|}{OAI} & \multicolumn{2}{c}{MIMIC-CXR} \\
        \cline{2-9}
        & PSNR$\uparrow$ & SSIM$\uparrow$ & PSNR$\uparrow$ & SSIM$\uparrow$ & PSNR$\uparrow$ & SSIM$\uparrow$ & PSNR$\uparrow$ & SSIM$\uparrow$ \\
        \hline
        Captured  & 16.68 & 91.35 & 24.76 & 89.85 & 17.71 & 89.84 & 19.27 & 83.79 \\
        UNet      & 21.85±0.16 & 90.71±0.32 & 25.64±0.07 & 91.09±0.07 & 17.79±0.24 & 85.95±0.45 & 23.25±0.44 & 86.92±0.19 \\
        MSEC      & 22.54±0.20 & \underline{92.25±0.12} & \textbf{30.46±0.17} & 91.36±0.10 & 19.62±0.21 & \underline{90.99±0.32} & \textbf{25.06±0.40} & 87.37±0.25 \\
        GRL       & 22.61±0.19 & 92.11±0.20 & 25.65±0.04 & 90.86±0.16 & 18.31±0.25 & 90.00±0.36 & 23.60±0.06 & 86.70±0.31 \\
        UFormer   & 22.57±0.35 & 91.06±0.08 & 25.98±0.23 & 89.57±0.74 & \underline{21.31±0.30} & 88.97±0.40 & 23.72±0.34 & 86.97±0.34 \\
        SwinIR    & 18.84±0.27 & 92.15±0.27 & 25.33±0.08 & 90.05±0.12 & 19.85±0.27 & 88.48±0.30 & 22.24±0.14 & 86.51±0.33 \\
        UHDFormer & \underline{22.85±0.24} & 91.76±0.19 & \underline{29.72±0.17} & \textbf{92.61±0.08} & 19.44±0.33 & \textbf{91.19±0.22} & 24.78±0.43 & \underline{88.27±0.41} \\
        Restormer & \textbf{23.09±0.05} & \textbf{92.33±0.11} & 28.55±0.09 & \underline{91.72±0.08} & \textbf{21.77±0.36} & 90.72±0.28 & \underline{25.05±0.49} & \textbf{88.46±0.39} \\
        \bottomrule
        \end{tabular}
\end{table*}


\begin{table*}[htbp]
    \renewcommand{\arraystretch}{1.25}
    \centering
    \caption{Diagnostic classification performance (AUC) on the MIMIC-CXR test dataset using Ark$^+$ diagnostic engine. Values are mean AUC with 95\% confidence intervals (subscript) from 1000 bootstrapped samples.}
    \begin{tabular}{l|c|c|c|c}
    \toprule
    \textbf{Condition} & \textbf{Captured} & \textbf{VisionCAD} & \textbf{Original} & \textbf{p-value} \\
    \hline
    Support Devices            & 0.835\textsubscript{(.810--.853)} & 0.901\textsubscript{(.884--.918)} & 0.925\textsubscript{(.907--.936)} & < 0.01 \\
    Pleural Effusion           & 0.853\textsubscript{(.830--.869)} & 0.890\textsubscript{(.871--.912)} & 0.894\textsubscript{(.881--.911)}& > 0.05 \\
    Lung Opacity               & 0.652\textsubscript{(.620--.679)} & 0.715\textsubscript{(.677--.741)} & 0.713\textsubscript{(.686--.740)}& > 0.05 \\
    Cardiomegaly               & 0.713\textsubscript{(.674--.738)} & 0.771\textsubscript{(.754--.788)} & 0.778\textsubscript{(.748--.802)}& > 0.05 \\
    Atelectasis                & 0.734\textsubscript{(.718--.763)} & 0.761\textsubscript{(.731--.778)} & 0.777\textsubscript{(.763--.797)}& < 0.05 \\
    Edema                      & 0.813\textsubscript{(.789--.845)} & 0.840\textsubscript{(.817--.856)} & 0.852\textsubscript{(.832--.869)}& < 0.01 \\
    No Finding                 & 0.755\textsubscript{(.732--.783)} & 0.809\textsubscript{(.781--.826)} & 0.814\textsubscript{(.790--.844)}& > 0.05 \\
    Pneumonia                  & 0.634\textsubscript{(.616--.648)} & 0.733\textsubscript{(.704--.758)} & 0.743\textsubscript{(.722--.762)}& > 0.05 \\
    Consolidation              & 0.678\textsubscript{(.645--.714)} & 0.725\textsubscript{(.683--.772)} & 0.729\textsubscript{(.692--.763)}& > 0.05 \\
    Enlarged Cardiomediastinum & 0.616\textsubscript{(.576--.665)} & 0.682\textsubscript{(.643--.714)} & 0.693\textsubscript{(.654--.746)}& > 0.05 \\
    Lung Lesion                & 0.704\textsubscript{(.628--.761)} & 0.765\textsubscript{(.733--.816)} & 0.773\textsubscript{(.708--.830)}& > 0.05 \\
    Pneumothorax               & 0.763\textsubscript{(.695--.816)} & 0.852\textsubscript{(.784--.905)} & 0.858\textsubscript{(.763--.924)}& > 0.05 \\
    Fracture                   & 0.654\textsubscript{(.554--.724)} & 0.731\textsubscript{(.660--.797)} & 0.763\textsubscript{(.684--.822)}& > 0.05 \\
    Pleural Other              & 0.795\textsubscript{(.727--.832)} & 0.860\textsubscript{(.822--.912)} & 0.872\textsubscript{(.841--.918)}& > 0.05 \\
    \hline
    \textbf{Average}           & \textbf{0.728} & \textbf{0.788} & \textbf{0.799} & - \\
    \bottomrule
    \end{tabular}
    \label{tab:Ark+_comparison}
\end{table*}

\begin{table*}[htbp]
    \renewcommand{\arraystretch}{1.2}
    \centering
    \caption{Comparison of automated report generation on the MIMIC-CXR dataset using state-of-the-art multimodal models. CheXpert $F_{1}$ scores are reported for 10 conditions. `Original` represents the original dataset images, while `VisionCAD` refers to the results from screen-captured images after being processed through our full VisionCAD pipeline. All values are multiplied by 100 for clarity. Subscripts on the VisionCAD columns indicate the performance change relative to the corresponding Original condition (\textcolor{darkgreen}{increase}, \textcolor{darkred}{decrease}).}
    \resizebox{\textwidth}{!}{
    \begin{tabular}{l|cc|cc|cc|cc}
    \toprule
    \multirow{3}{*}{Condition} & \multicolumn{4}{c|}{SOTA VLM Model} & \multicolumn{4}{c}{SOTA VLM Thinking Model} \\
    \cline{2-9}
    & \multicolumn{2}{c|}{Original} & \multicolumn{2}{c|}{VisionCAD} & \multicolumn{2}{c|}{Original} & \multicolumn{2}{c}{VisionCAD} \\
    & w/o Ark$^{+}$ \ding{108} & w/ Ark$^{+}$ \ding{110} & w/o Ark$^{+}$ \ding{108} & w/ Ark$^{+}$ \ding{110} & w/o Ark$^{+}$ \ding{115} & w/ Ark$^{+}$ \ding{117} & w/o Ark$^{+}$ \ding{115} & w/ Ark$^{+}$ \ding{117} \\
    \hline
    Support Devices           & 55.71 & 18.65 & 48.94$_{\textcolor{darkred}{-6.77}}$  & 32.11$_{\textcolor{darkgreen}{+13.46}}$ & 61.25 & 66.19 & 62.50$_{\textcolor{darkgreen}{+1.25}}$ & 64.63$_{\textcolor{darkred}{-1.56}}$\\
    Pleural Effusion          & 33.75 & 69.32 & 30.22$_{\textcolor{darkred}{-3.53}}$  & 67.71$_{\textcolor{darkred}{-1.61}}$  & 54.91 & 65.47 & 54.34$_{\textcolor{darkred}{-0.57}}$ & 65.17$_{\textcolor{darkred}{-0.30}}$  \\
    Lung Opacity              & 40.00 & 43.93 & 45.89$_{\textcolor{darkgreen}{+5.89}}$  & 46.92$_{\textcolor{darkgreen}{+2.99}}$  & 50.10 & 47.91 & 52.41$_{\textcolor{darkgreen}{+2.31}}$ & 50.77$_{\textcolor{darkgreen}{+2.86}}$  \\
    Cardiomegaly             & 41.64 & 46.33 & 40.44$_{\textcolor{darkred}{-1.20}}$  & 42.91$_{\textcolor{darkred}{-3.42}}$  & 48.51 & 55.38 & 46.23$_{\textcolor{darkred}{-2.28}}$ & 53.93$_{\textcolor{darkred}{-1.45}}$  \\
    Atelectasis              & 7.41  & 48.57 & 7.94$_{\textcolor{darkgreen}{+0.53}}$   & 44.57$_{\textcolor{darkred}{-4.00}}$  & 28.87 & 48.09 & 27.05$_{\textcolor{darkred}{-1.82}}$ & 47.06$_{\textcolor{darkred}{-1.03}}$  \\
    Edema                   & 31.62 & 51.08 & 34.12$_{\textcolor{darkgreen}{+2.50}}$  & 46.98$_{\textcolor{darkred}{-4.10}}$  & 41.82 & 48.61 & 38.59$_{\textcolor{darkred}{-3.23}}$ & 50.53$_{\textcolor{darkgreen}{+1.92}}$  \\
    No Finding              & 29.81 & 27.03 & 34.09$_{\textcolor{darkgreen}{+4.28}}$  & 20.62$_{\textcolor{darkred}{-6.41}}$  & 10.64 & 31.40 & 8.59$_{\textcolor{darkred}{-2.05}}$  & 37.16$_{\textcolor{darkgreen}{+6.96}}$ \\
    Pneumonia               & 17.91 & 19.05 & 9.23$_{\textcolor{darkred}{-8.68}}$   & 18.75$_{\textcolor{darkred}{-0.30}}$  & 16.67 & 18.92 & 15.27$_{\textcolor{darkred}{-1.40}}$ & 17.65$_{\textcolor{darkred}{-1.27}}$ \\
    Consolidation           & 3.70  & 14.29 & 4.08$_{\textcolor{darkgreen}{+0.38}}$   & 14.17$_{\textcolor{darkred}{-0.12}}$  & 17.39 & 19.51 & 14.49$_{\textcolor{darkred}{-2.90}}$ & 21.22$_{\textcolor{darkgreen}{+1.71}}$ \\
    Enlarged Cardiomediastinum & 6.67  & 18.18 & 7.69$_{\textcolor{darkgreen}{+1.02}}$   & 7.14$_{\textcolor{darkred}{-11.04}}$ & 10.26 & 12.66 & 7.61$_{\textcolor{darkred}{-2.65}}$ & 4.82$_{\textcolor{darkred}{-7.84}}$ \\
    \hline
    CheXpert $F_{1}$ (avg.) & 26.82 & 35.64 & 26.26$_{\textcolor{darkred}{-0.56}}$ & 34.19$_{\textcolor{darkred}{-1.45}}$ & 34.04 & 41.41 & 32.71$_{\textcolor{darkred}{-1.33}}$ & 41.29$_{\textcolor{darkred}{-0.12}}$ \\
    \bottomrule
    \end{tabular}
    }
    \label{tab:comparison_of_report_accurancy}
    \vspace{10pt}
    \raggedright
    \textit{Wilcoxon signed-rank test indicates no statistically significant difference between Original and VisionCAD results within each model configuration (e.g., comparing Original w/o Ark$^{+}$ \ding{108} with VisionCAD w/o Ark$^{+}$ \ding{108}). All such paired comparisons have $p > 0.05$.}
\end{table*}

\end{document}